\documentclass{article}
\usepackage{amsmath} 
\usepackage{amssymb}  
\usepackage{graphicx}
\graphicspath{ {./images/} }
\usepackage{tabularx}
\usepackage{makecell}
\usepackage[compact]{titlesec}
\titlespacing{\section}{0pt}{2ex}{1ex}
\titlespacing{\subsection}{0pt}{1ex}{0ex}
\titlespacing{\subsubsection}{0pt}{0.5ex}{0ex}
\usepackage{authblk}
\usepackage[ruled,vlined]{algorithm2e}

\usepackage[preprint]{corl_2020} 

\title{Towards Autonomous Eye Surgery by Combining Deep Imitation Learning with Optimal Control}

\author[1]{Ji Woong Kim}
\author[1]{Peiyao Zhang}
\author[2]{Peter Gehlbach}
\author[1]{Iulian Iordachita}
\author[1]{Marin Kobilarov}
\affil[1]{Department of Mechanical Engineering, Johns Hopkins University}
\affil[2]{Wilmer Eye Institute, Johns Hopkins University School of Medicine}
\affil[ ]{{\{jkim447, pzhang24, iordachita, marin\}@jhu.edu} and pgelbach@jhmi.edu}

\begin{document}
\maketitle

\vspace{-24pt}
\begin{abstract}
    During retinal microsurgery, precise manipulation of the delicate retinal tissue is required for positive surgical outcome. However, accurate manipulation and navigation of surgical tools remain difficult due to a constrained workspace and the top-down view during the surgery, which limits the surgeon's ability to estimate depth. To alleviate such difficulty, we propose to automate the tool-navigation task by learning to predict relative goal position on the retinal surface from the current tool-tip position. Given an estimated target on the retina, we generate an optimal trajectory leading to the predicted goal while imposing safety-related physical constraints aimed to minimize tissue damage. As an extended task, we generate goal predictions to various points across the retina to localize eye geometry and further generate safe trajectories within the estimated confines. Through experiments in both simulation and with several eye phantoms, we demonstrate that our framework can permit navigation to various points on the retina within 0.089mm and 0.118mm in xy error which is less than the human's surgeon mean tremor at the tool-tip of 0.180mm. All safety constraints were fulfilled and the algorithm was robust to previously unseen eyes as well as unseen objects in the scene. Live video demonstration is available here: \url{https://youtu.be/n5j5jCCelXk}  
\end{abstract}

\keywords{deep imitation learning, optimal control, eye surgery, ophthalmology} 


\begin{figure}[h]
        \centering
        \includegraphics[width=1.0\textwidth]{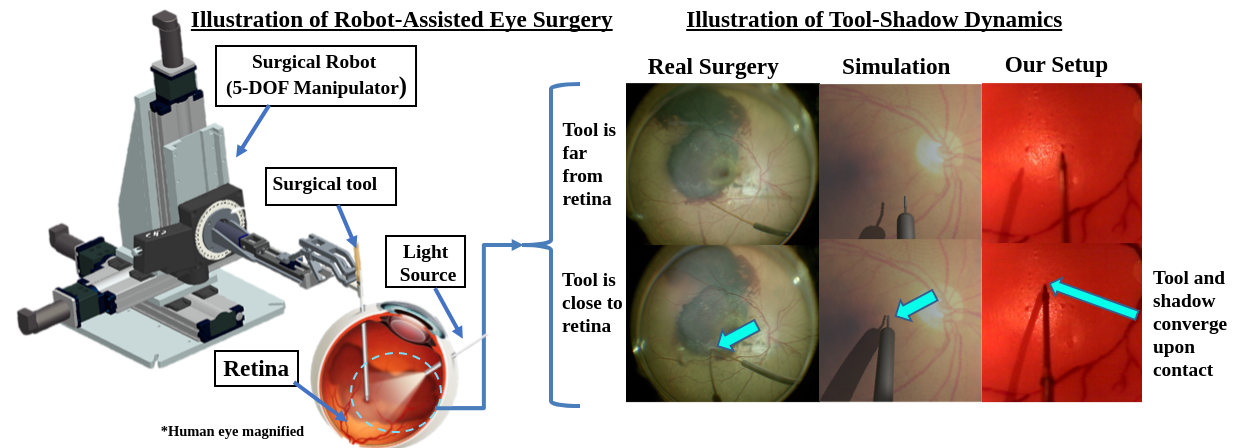}
        \caption{\small  (Left) the surgical tool is inserted into the human eye through a sclera port and the light source projects a shadow which can be used as cues to estimate depth (right) demonstration of tool-shadow dynamics}
        \label{fig:intro_pic}
        \vspace{-12pt}
\end{figure}

\section{Introduction}
\vspace{-7pt}	
  Retinal surgery is among the most challenging of microsurgical procedures requiring precise manipulation of the delicate retinal tissue. The surgery proceeds with the surgeon inserting a surgical tool through a sclera port and visualizing the retina and the surgical tools from a top-down view using an operating microscope. During surgery, one of the most challenging tasks is precisely navigating the surgical tool-tip to the desired position on the retinal tissue. For example, while performing retinal vein cannulation or retinal membrane peeling, the surgeon must intuitively localize the tool-tip with respect to the retina in order to land the surgical tool-tip precisely at the desired tissue location. Several factors contribute to its difficulty, such as viewing the surgery from top-down view which can hinder the surgeon's depth-estimation accuracy, visual distortions caused by the human lens, and the limited depth of field of the operating microscope. Precise navigation of surgical tools is required for most retinal procedures, yet it remains among the more challenging tasks, requiring years of experience to master.

\begin{figure}[h]
        \centering
        \includegraphics[width=1.0\textwidth]{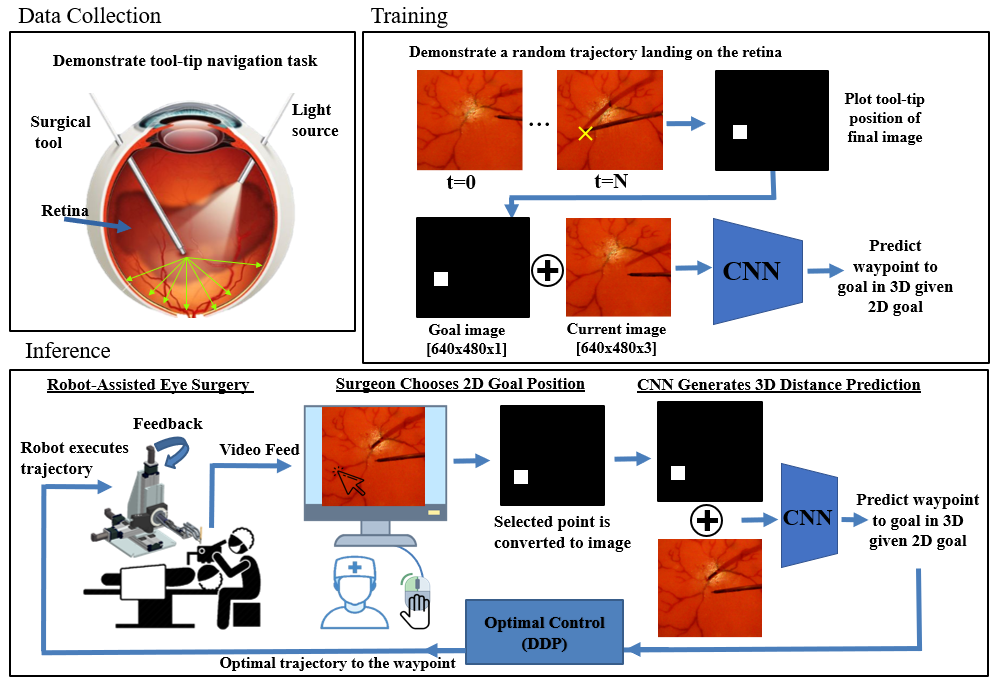}
        \caption{\small Data collection, training and inference process for achieving autonomous navigation setup}\vspace{-5pt}
        \label{fig:system}
        \vspace{-5pt}
\end{figure}


In this work, we propose to automate the tool navigation task by learning to predict relative goal positions from video and by generating smooth trajectories to these goals while satisfying known physical kinematic constraints. The learned goal-prediction network estimates distance and direction from the tool-tip to various target locations on the retinal surface, in a manner similar to how surgeons utilize visual cues to safely navigate the tool-tip with respect to the retina. Specifically, the input to our proposed network is a top-down monocular view of the surgery and user-input defining the 2-d goal location to be reached i.e. as simple as clicking the desired location using a mouse Fig.\ref{fig:system}. Given these inputs, the output of the network is a 3-d vector (direction and distance) to the clicked position. The advantage of this approach is that the user specifies the goal in 2-d, and the network outputs a prediction in 3-d space; since estimating depth is the challenging task for humans, the network takes the burden by predicting distance along the depth dimension based on its training experience. The further advantage of this learning setup is that with goal predictions across various points on the retina, the retinal geometry can be reconstructed and employed for safe tool navigation. We include a demonstration of this approach, and as an extended implementation, we demonstrate a blood-vessel following task where the surgical tool-tip hovers over the localized retinal geometry while staying as close as ~0.200$\mu m$ to the retinal surface. This enables safe planning of various trajectories inside the eye.



Another important consideration in achieving autonomous navigation is the physical constraint posed by the sclera point, i.e. the hole through which the surgical tool is inserted into the eye. The sclera point can be considered as a point constraint which should not be violated in lateral directions; large deviations form the sclera point could cause eye rotation and thus eye-muscle damage. To address this issue, we integrate an optimal control framework that enables path-planning of trajectories to the retina while respecting the sclera constraint. Specifically, we utilize differential dynamic programming (DDP) \cite{ddp}, which is a numerical optimization approach to path-planning based on specified costs. To satisfy the point constraint of the sclera, we simply add a cost to minimize the distance between the sclera point and the tool-axis when searching for an optimal trajectory. In extended tasks such as blood-vessel following, we utilize this framework to constrain the trajectories to remain close to and without colliding with the retinal surface. Ultimately, our system combines both a learning-based approach and optimal control framework, where the goal-prediction network generates a goal waypoint on the retina and the DDP algorithm plans an optimal trajectory to that goal. During execution of the perceptual inference (including goal prediction and eye geometry reconstruction), the optimal trajectory is repeatedly re-computed at some fixed frequency, e.g. 5 Hz, until the tool-tip reaches its target.

We also note that our approach is based on the assumption that learning autonomous navigation is possible primarily using vision. In fact, surgeons rely on their visual perception to localize objects and spatially estimate their proximity to perform retinal surgery. Furthermore, the tool-shadow dynamics provide an important cue that helps with detecting proximity between the tool-tip and the retina. Specifically, when the tool approaches the retina, the tool and its shadow converge, which can be used as cues to train the network (Fig.\ref{fig:intro_pic}).  

To assess the accuracy of our system, we employ a benchmark task where the surgical tool-tip must reach a predefined array of positions on the retina. The objective of this experiment is to assess how well the network can navigate to various positions on the retina given a specific goal image. As an extension of this task, we also consider localization of the retina by predicting distances to various points across the retinal surface and reconstructing the eye-geometry. Furthermore, using the localized eye, we demonstrate the usefulness of the localization by demonstrating a blood-vessel following task, which can be challenging even for the surgeons when achieving constant micron-scale off-set is the goal.
	

\begin{figure}[h]
        \centering
        \includegraphics[width=1.0\textwidth]{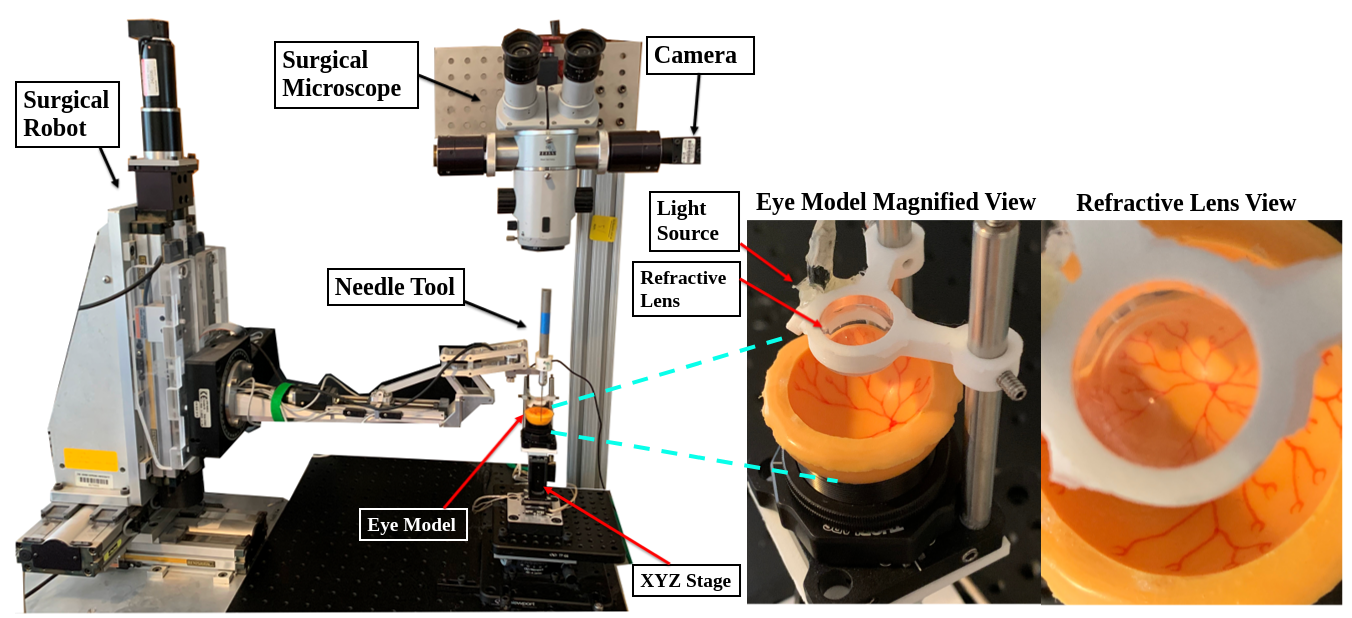}
        \caption{\small Phantom experiment setup}\vspace{-5pt}
        \label{fig:exp_setup}
        \vspace{-5pt}
\end{figure} 


\section{Related Work}
\vspace{-7pt}
\subsection{Robotic Eye Surgery}
Past works in robotic eye surgery have focused primarily on state estimation or detection system that assist the surgeon by providing extra information. \citet{japan_shadow} modeled a proximity detection system that forecast detection when the tool-tip and the shadow-tip converged by a pre-defined pixel distance. ~\citet{6212340} applied stereo vision setup that estimated the depth of the tool and the retina respectively thereby demonstrating a proximity detection system. Optical coherence tomography (OCT) has also been integrated into both the surgical microscope and robotic tools. OCT is a laser imaging modality that gives distance readings from the optical fiber tip to the retina with micron-scale accuracy \citep{4d_OCT} \citep{ioct}. However, due to its slow processing speed and high-cost, its use in the operating room has been limited. Contrary to laser-imaging based depth predictions, some works have attempted to use stereo imaging to reason about depth including \citep{retinal_3d_reconstruction_van_gool} and ~\citep{6212340}. However, their studies have been limited to open-sky phantoms (i.e. no refractive lens between the retina and the microscope) and dealing with unknown distortions and out-of-focus images pose a significant challenge for stereo-based methods. Ultrasound imaging, which is another alternative, has generally lacked the necessary precision for retinal tool navigation but is useful for certain anterior segment applications.

\subsection{Learning}
Many works have demonstrated the effectiveness of deep learning in sensorimotor problems, such as in playing computer games \citep{atari_games, learn_to_act} and vision-based navigation  \citep{nav_1, drive_2, drive_3, MIT_autonomous_driving, conditional_learning, chauffeur_net}. Our network design borrows from an approach proposed in \citep{MIT_autonomous_driving} and \citep{chauffeur_net}, which uses a user-defined goal represented as topographical binary image. Similarly, in our approach for designing the network, we include a topographical binary image defining the mouse-click position as input to a network. Directly relevant to eye surgery, recent works have utilized OCT imaging and reinforcement learning to demonstrate needle insertion using tool-pose states as input \citep{keller2020optical}, although vision is not used. More recently, a vision-based autonomous navigation system was developed for retinal surgery, though without the inclusion of optimal control to impose appropriate physical constraints \citep{autonomous_navigation_retina}. Furthermore, our approach considers predicting goal point to a surrounding object (i.e. the retina) rather than predicting an incremental waypoint to the goal, which enables further capabilities such as localization of the retina itself.   

\section{Problem Formulation}
\vspace{-7pt}
Consider  a robotic manipulator with end-effector (i.e. surgical needle) state defined by $x=(p,R,v,\omega)$ where $p\in\mathbb{R}^3$ defines the tool-tip position, $R\in SO(3)$ the orientation matrix, $v\in\mathbb{R}^3$ the base-frame translational velocity and $\omega\in\mathbb{R}^3$ the end-effector-frame angular velocity. The end-effector pose is defined as $q=(p,R)\in SE(3)$. The robot is fully actuated and for our purposes the control inputs are the forces $u=(u_v, u_\omega)$ defining the translational forces $u_v\in\mathbb{R}^3$ and torques $u_\omega\in\mathbb{R}^3$ in end-effector frame (the robot internal controller can map those to appropriate joint-angle torques). Let the robot end-effector occupy a region $A(q)\subset \mathcal W$ in the workspace $\mathcal W\subset \mathbb{R}^3$.


The end-effector state $x$ is fully observable using high-precision joint encoders and precise knowledge of the robot forward kinematics. In addition, the system is equipped with workspace-fixed microscope video camera generating observations $o(t)\in \mathcal I$ from the space of images $\mathcal I$ capturing the surgical scene. 
 
The procedure begins when the surgeon inserts the tool into the sclera. The sclera entry point $p_s\in\mathbb{R}^3$ is recorded by the robot and must remain fixed after each entry during surgery.~\footnote{this point is often called \emph{remote center of rotation} (RCM) as it acts as a fixed point through which the instrument can rotate but only translate "longitudinally" thus avoiding lateral forces that could damage the sclera.} The surgeon then selects a sequence of $N_g$ pixels (i.e. 2-d points in image space) denoted by $g_i \in o$ for $i=1,...,N_g$, that correspond to a set of euclidean points (i.e. 3-d points in robot frame) on the surface of the retina inside the eye to which the robot must navigate, denoted by $p_i\in E$, where $E\subset \mathcal W$ defines the surface of the eye retina, i.e. the boundary of the eye internal geometry.

The objective is to control the robot to safely navigate to the specified goal locations. More formally, we seek to generate an end-effector trajectory $x([t_0,t_f])$ over some time-interval $[t_0,t_f]$ defined as:
\begin{align}
    &  \arg\!\min_{\!\!\!\!\!\!\!\! x(\cdot),u(\cdot)} \int_{t_0}^{t_f} C(x(t), u(t)) dt, &&\quad \text{:minimize cost, subject to:} \label{eq:problem_cost} \\
    & \dot x(t) = f(x(t), u(t)), &&\quad \text{:robot-tissue dynamics} \label{eq:problem_dynamics}\\
    & (I+r_x(t) r_x(t)^T)(p_s - p(t)) = 0,  &&\quad \text{:sclera constraint} \label{eq:problem_sclera}\\
    & \mathcal A(q(t)) \cap E = \emptyset, &&\quad \text{:collision constraint} \label{eq:problem_collision}\\
    & p(t_i) = p(t_0) + R(t_0)F(o(t_0), g_i), \text{ for } i=1,\dots, N_g, &&\quad\text{:visit goal locations} \label{eq:problem_goals}\\
    & x(t_f) \in X_f,  &&\quad \text{:terminal constraint} \label{eq:problem_terminal}
\end{align}
where $C(x,u)$ is a given cost function e.g. ensuring smooth and safe motion, and $r_x= Re_x$ with $e_x=(1,0,0)$ being the first basis vector. The function $F(o, g)$ is the mapping from an image $o$ and a selected goal pixel $g\in o$ at time $t_0$ to a 3-d position in end-effector-frame corresponding to the exact goal location on the retinal surface. Note that neither the eye geometry $E$ nor the mapping $F$ are available a priori and must be constructed from available sensor data.

We consider the following two problems:
\begin{enumerate}
\item \emph{Reaching a given goal point}: safely reach a single goal location, i.e. $N_g=1$, corresponding to e.g. a desired retinal vein puncture point that we reach with zero velocity, i.e. $x(t_f)\in X_f$ simply corresponds to $v(t_f)=0$ and $\omega(t_f)=0$.

\item \emph{Vessel following}: generate a smooth and safe navigation through a sequence of given goal image locations, where the robot does not need to stop at each point, i.e. $N_g>1$ while the terminal zero-velocity constraint is applied to the last goal point only. 
\end{enumerate}
\begin{figure}[h]
        \centering
        \includegraphics[width=0.7\textwidth]{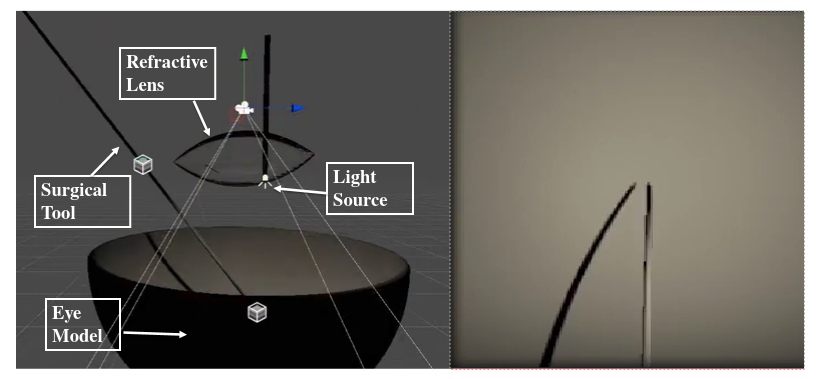}
        \caption{\small Simulation setup; (left) side view (right) camera view navigation setup}\vspace{-5pt}
        \label{fig:sim}
        \vspace{-5pt}
\end{figure}


\section{Technical Approach}
\vspace{-7pt}
To solve~\eqref{eq:problem_cost}--\eqref{eq:problem_terminal} we first address the computation of the relative goal point map $F(o, g)$ which is accomplished through machine learning using prior demonstrations with labeled contact points. The eye geometry $E$ is then constructed by fusing multiple pixel position predictions (from the neural network) through a least-squares approach to estimate the local euclidean shape assuming locally spherical geometry. These two components are then used in a discrete-time numerical optimal control framework that additionally enforces sclera and collision constraints. 

\subsection{Learning to Predict Goal Point} \label{predict_distance}
We formulate the problem as a goal-based imitation learning setup, where the objective is to estimate the required motion to place the surgical tool-tip precisely at the user-specified goal position on the retina, similar to how surgeons reason about moving the tool-tip to retina. We denote this relative motion by the vector $d\in\mathbb{R}^3$, which encodes both the \emph{distance} to the goal $\|d\|$ and also the unit vector $\frac{d}{\|d\|}$ encoding the \emph{direction} of motion.  Specifically, given a data-set of tool trajectories $D=\{(o_i, g_i), d_i\}_{i=1}^{N}$, the objective is to construct a function approximator $d=F((o, g);\theta)$ with parameters \(\theta\), that maps observation-goal pair to vector-to-goal generated by the expert. The objective function can be expressed as the following: 

\begin{gather}
\arg\min_{\theta} \sum_{i=1}^{N} L(d_i, F(o_i, g_i; \theta)),
\end{gather}

where $L$ is a given loss function. In our work, we choose the observation to be a monocular image $o\in \mathcal I$ of the surgical scene viewed top-down via a microscope, the goal input to be $g_i=(x_i,y_i)\in\mathbb{R}^2$ which specifies the desired projected 2-d position on the retinal surface, corresponding to vector-to-goal $d$. The ground truth goal-prediction is generated via robot kinematics.
\subsection{Localizing the Retina} \label{localize_retina}
Using goal predictions across various points on the retina as described in \ref{predict_distance}, we fit a sphere to these points by fitting a radius and center point using least-squares method. Specifically, we algebraically rearrange the terms of the  sphere equation, $\|p-p_0\|^2=r^{2}$, to obtain as the linear relationship $\vec{f}=A \vec{c}$ with:
\begin{gather}
    \vec{f}=\left[\begin{array}{c}
    \|p_i\|^2\\
    \|p_{i+1}\|^2 \\
    \vdots \\
    \|p_n\|^2
    \end{array}\right], \quad 
    A=\left[\begin{array}{cccc}
    2 p_i^T & 1 \\
    2 p_{i+1}^T & 1 \\
    \vdots & \vdots \\
    2 p_n^T & 1
    \end{array}\right] ,\quad
    \vec{c}=\left[\begin{array}{c}
    p_{0} \\
    r^{2}-\|p_0\|^2
    \end{array}\right],
\end{gather}

where we solve for $\vec{c}$. In the equation, $ \{ p_i, ..., p_n\} \in\mathcal W\subset \mathbb{R}^3$,  are the estimated points on the retinal surface obtained using the goal-prediction network, and $p_0$ and $r$ is the estimated center position and radius of the fitted sphere. The least-squares method is combined with random sample consensus (RANSAC) to search for the solution with the highest number of inliers, based on a prescribed inlier threshold.
\vspace{-5pt}
\subsection{Optimal Control Formulation}
In order to reach a waypoint generated by the network, we must navigate within physical constraints to avoid damage to the eye. To accomplish this reliably in real-time we re-formulate the optimal control problem to include the constraints~\eqref{eq:problem_sclera}--\eqref{eq:problem_collision} as least-squared penalties, according to:
\begin{align}\label{opt_ctrl}
\begin{split}
C=\frac{1}{2} x\left(t_{f}\right)^T P_{f} x\left(t_{f}\right) +\int_{t_{0}}^{t_{f}} &\frac{1}{2}\left[u(t)^{T} R(t) u(t)\right]  \\ 
& + w_s \cdot \| (I+r_x(t) r_x(t)^T)(p_s - p(t))\|^2  \\
& + w_e \cdot \delta_{\{\|d_c(t)\| > r-\epsilon\}} \left\|d_c(t) -  r \frac{d_c(t)}{\|d_c(t)\|} \right\|^2 \mathrm{d}t,
\end{split}
\end{align}

where $p_c \in\mathbb{R}^3$  is the center, $d_c(t)=x(t)-p_c$, and $r \in\mathbb{R}^3$ is the radius of the estimated sphere obtained using the least-squares approach described in \ref{localize_retina}. In general the cost aims to minimize the error in reaching the goal (encoded using $P_f\geq 0$ gain matrix), control effort (using $R>0$ gain matrix), penalize deviation from sclera points (using weight $w_s$), and penalize collision with the locally spherical retinal surface if the tool is within some $\epsilon>0$ from it (using weight $w_e$), encoded by the delta function $\delta_{\{\|d_c(t)\| > r-\epsilon\}}$. The optimal control problem is solved numerically using differential dynamic programming (DDP) based on a discrete-time quantization of the robot motion using some fixed time-step $dt$. For autonomous navigation to a single goal point, we omit the non-penetration constraint to the retina. For vessel following we include both sclera the non-penetration constraint in order navigate along the blood-vessels while staying close to the retina. We note that with this re-formulation, the hard constraints are now enforced as soft constraints. In practice, the constraints can still be ensured by setting the gains of the safety constraints to be significantly higher than the gains of the smoothness and the goal-error terms. Furthermore, this re-formulation enables increased numerical stability and efficiency.


\begin{figure}[h]
        \centering
        \includegraphics[width=1.0\textwidth]{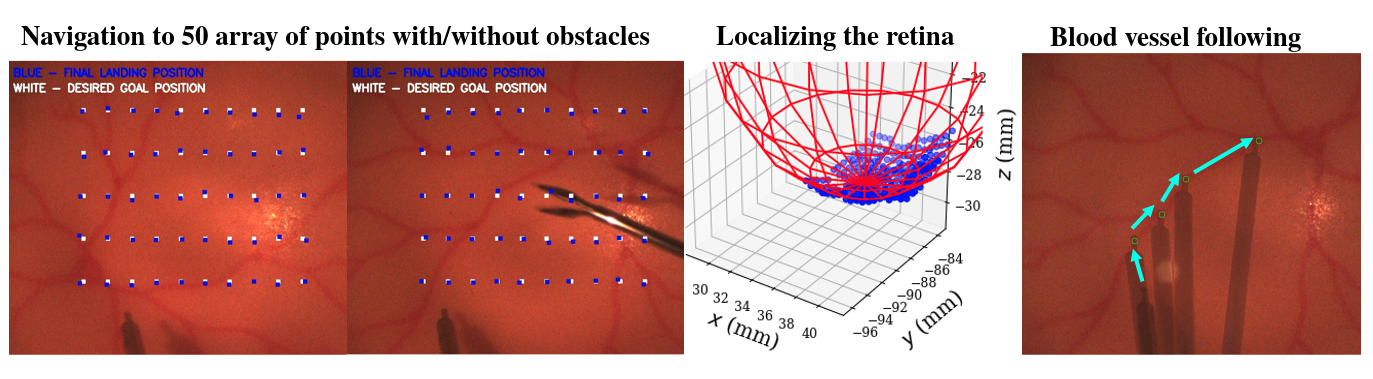}
        \caption{\small Results on autonomous navigation, eye localization, and vessel following tasks}\vspace{-5pt}
        \label{fig:task_demo}
        \vspace{-5pt}
\end{figure}

\section{Method}
\vspace{-7pt}
\subsection{Eye Phantom and Simulation Experimental Setup}
Our setup consists of a surgical robot, rubber eye model,  microscope, surgical needle tool, and light source (chandelier) as shown in Fig.\ref{fig:exp_setup}. The rubber eye is mounted on a piezo-electric XYZ stage in order to move the eye to simulate involuntary movements of the patient. We also mount a refractive lens between the eye model and the microscope to simulate the distortions encountered during real surgery e.g. the human lens and the vitreous. The dimension of the eye model we used is 25.4mm, which is slightly larger than the average human eye size, which ranges from 20 - 22.4mm \cite{eye_diameters}. There are however 25.4 mm human eyes in myopic individuals. For sense of scale, the width of the surgical tool-tip measures 0.32mm, and the operation space is ~12x16mm (Fig.\ref{fig:task_demo}).

We also demonstrate our setup in a simulated environment using Unity3D. We created an environment similar to our phantom setup by including refractive lens in to the scene. Furthermore, we explored a more involved domain randomization by changing the orientation and the position of the light source, and changing the orientation and position of the refractive lens. The benefit of using the simulator is that it gives an approximate upper-bound performance we can expect from phantom experiments, since the environment parameters can be perfectly controlled without errors. 

\subsection{Data Collection}
The data collection procedure consists of navigating the surgical tool-tip to various locations on the retinal surface. We first estimate the retinal geometry by sampling its surface with the surgical tool and fitting a sphere to the sampled points using the least squares method (\ref{localize_retina}). Once the retinal surface is localized, we demonstrate many tool navigation trajectories across the retinal surface starting from various initial configurations of the tool. For simplicity, for each collected trajectory, we navigate to a single targeted location in a straight-line trajectory. Since the retinal surface had been localized, the tool-tip makes very gentle contact with the retina. Throughout the trajectory, images from the microscope and tool-tip positions based on robot kinematics are saved. After each trajectory demonstration, the piezo-stage was moved $\pm0.30$mm in XYZ directions for domain randomization. We also used two different eyes at varying orientation to generalize to changing eye textures. The pseudocode for data collection is summarized in Appendix \ref{appendix_pseudocode}.

\subsection{Network Training}
To train the goal-prediction network, we chose Resnet-18.  The input of the network are current surgical image (640x480x3) and the goal image (640x480x1) stacked along the channel dimension. The goal image marks the desired landing position of the tool-tip on the retina. Specifically, given $n$ frames $I_1, ... I_n \in \mathcal I$, $n$ vector-to-goal values $d_1,\dots, d_n \in \mathbb{R}^3$, and a goal image coordinate $g \in \mathbb{R}^2$ of this trajectory, the input-output relationship can be expressed as $(\text{input}, \text{output})=\left((I_t, g), d_{t}\right)$, for $t=1,\dots,n$. We also took the approach of discretizing the output space of the network, as it proved to learn better than a continuous output case. Specifically, we discretized the continuous x, y, z coordinates into 580, 1345, and 320 bins.  We set the optimizer to be the Adam optimizer~\cite{kingma2014adam}. The initial learning rate and batch size were set as 0.0003 and 128. We express the loss functions as
\begin{align}
    L(b, \hat p) = \sum_{j\in \{x,y,z\}} \sum_{c=1}^{M_j} -b_{j,c} \log(\hat p_{j,c}),
\end{align}
where $b_{j,c}$ are binary indicators for the true class label $c$, and $\hat p_{j,c}$ are the predicted probability that the coordinate $j$ is of class $c$. Errors from all three dimensions $j\in \{x,y,z\}$ have been combined into one cost function. As specified above, we employed $M_x=580,\ M_y=1345,\ M_z=320$ bins.

\section{Results and Discussions}
\vspace{-7pt}
To assess the accuracy of our results, we implemented a benchmark task where the network must navigate the tool to fifty predefined locations on the retina as shown in Fig. \ref{fig:task_demo}. The objective of this task is to assess how well and the network can navigate to various points on the retina.  For our second task, we use the goal-prediction network to sample an array of points across the retinal surface to obtain many distance readings to the retina. These points were then used to reconstruct the eye surface using the least squares approach. To confirm our solution, we used the estimated retinal geometry to navigate to the phantom retina at random and checked from side view whether it was making gentle contact. Also, the estimated geometry was compared to a solution achieved by manually sampling the eye with the tool-tip, which we regarded as the ground-truth. For our last task, we used the estimated geometry of the eye to follow a specific segment of a blood vessel while hovering above to avoid collision. The result of these tasks are shown in Fig. \ref{fig:task_demo}.

Our numeric results are summarized in Table \ref{tab:exp_results}. In the first column of Table \ref{tab:exp_results}, we demonstrate that we can navigate to various positions on the retina with an average xy accuracy of 0.072mm and 0.130mm respectively. The visual result is shown in the leftmost figure of Fig \ref{fig:task_demo}. The white squares denote the fifty array of positions to visit and the blue squares denote the final landing position of the tool-tip navigated by the network. For sense of scale, the tool-tip width measures 0.32mm. In the second column, we demonstrate that our navigation system is robust to the presence of an unseen object. The unseen distractor object is a forcep, which is a commonly used surgical tool. Throughout testing, We did not observe a visual difference in performance, and thus the average error was similar to without-distraction scenario at 0.089mm and 0.118mm in xy error. The visual result of this test is shown in the second image in Fig.\ref{fig:task_demo}. For our second task of localizing the eye model, we demonstrate an error of 0.53mm, 0.084mm and 0.292mm in estimating the center location. Specifically, the ground-truth center of the eye was located at (39.26, -89.58, -14.32)mm and our estimated center was estimated at (38.73,	-89.66, -14.61)mm location. To confirm whether the estimated center was reasonable, we sampled the eye 40 times at random using the estimated center and radius and visually confirmed that the tool-tip gently touched phantom retina every time. Using this estimated retinal geometry, we applied it to a vessel-following task to follow a trajectory along a chosen blood-vessel. To ensure that no collision occurs, we decreased radius of the estimated sphere by 0.2mm, and an appropriate trajectory was generated by the optimal control framework above the retinal surface at 0.2mm height. The trajectory-tracking error of vessel following is reported in the last column of Table \ref{tab:exp_results}. Furthermore, the sclera error of all task is shown on the fourth row of Table \ref{tab:exp_results}. No trajectories exceeded a mean sclera error of 0.704mm across all tasks, satisfying required constraints for safe surgery. We also note that eye geometry estimation and vessel following tasks require short and easy procedures. Specifically, eye geometry estimation may take less than a second as it simply requires multiple surgical images. Vessel-following may require 2-3 seconds as it requires the surgeon to click the desired waypoints of interest using a mouse. 

For our simulation results, we only demonstrated the autonomous navigation task. The overall combined xy error in reaching 50 predefined locations was 0.119mm, which is significantly less than the errors encountered in the phantom experiments. This is expected since simulation environment states and collision detection are perfect. Still, given the fact the simulation involved more substantial randomization in lens movement and light source movement, it implies similar sophistication can be learned in the real-world and potentially in real surgery (Appendix \ref{appendix_sim}).
\vspace{-12pt}


\begin{table}[]
\caption{\label{tab:exp_results} Error on Various Tasks}
\centering
\begin{tabular}{c|cccc}
\hline
\thead{Error \\category}            & \thead{Navigation to \\retina (50 positions)\\ (mm, pixels)}                            & 
             \thead{Navigation to retina \\ w/ forcep distraction \\ (50 positions) \\ (mm, pixels)}                          & 
             \thead{Eye localization \\center, radius \\(mm)}     & 
             \thead{Vessel-following \\ trajectory \\tracking (mm)}                                  \\
             \hline
x error       & 0.072, 1.8 & 0.089, 2.24                 & 0.53  & 0.002    \\
y error & 0.130, 3.26 & 0.118, 2.94 & 0.084 & 0.003 \\
z error       & NA                           & NA                         & 0.292 & 0.006    \\
sclera error & 0.704mm                        & 0.613mm & NA    & 0.059 \\
radius        & NA                           & NA                         & 0.152 & NA                                
\end{tabular}
\end{table}
\section{Conclusion}
\label{sec:conclusion}
\vspace{-7pt}
In this work, we demonstrate that autonomous navigation is possible by learning to predict goal positions on the retina from user-specified image pixel locations. Our work has been tested in simulated eyes as well as eye phantom models where eye movement, different textures, refractive view, and various lighting conditions are encountered. We demonstrate that through a benchmark experiment, our method is capable of achieving higher accuracy in navigating to a target tissue than a surgeon, whose error is characterized by mean-amplitude of hand-tremor at the tool-tip. Still, more work is necessary to adopt our work to real human surgeries. In addition, human eyes have a slightly elliptical shape and the proposed spherical regression must be generalized accordingly. Finally, rigorous experiments with real physical tissue are critical to enable to application of the proposed techniques in a clinical setting. To this end, we are beginning to employ porcine eyes, which share a similar anatomy as human eyes.



\clearpage


\bibliography{bib}  

\newpage
\renewcommand{\thesection}{\Alph{section}} 
\setcounter{section}{0} 

\section{Pseudocode for Data Collection and Inference}
\label{appendix_pseudocode}
\begin{algorithm}[H]
\SetAlgoLined
 sample ~30 points on the eye phantom\;
 center, radius = leastSquares(sampled points)\;
 \While{iter $<$ n}{
  choose random initial tool-tip point and random sclera point\;
  goal point = generateRandomGoalOnRetina(center, radius)\;
  randomize xyz position of the eye\;
  \While{goal point not reached}{
   moveToward(goal point)\;
   record image and robot kinematics\;
  }
}
\caption{Data Collection}
\end{algorithm}

\begin{algorithm}[H]
\SetAlgoLined
 \For{each [initial point, goal point]}{
   \While{numIter $<$ maxIter}{
    waypoint = CNN(goal point, current image)\;
   trajPoints, trajVelocity = optimalControl(waypoint, current tool-tip position)\;
       \If{z-distance to goal $>$ 0.1mm}{
         closedLoopControlToReachGoal(trajPoints, trajVelocity)\;
        }
   } 
}

\caption{Inference}
\end{algorithm}

\section{Error Calculation for the Benchmark Task}
For the benchmark task of navigating to fifty points across the retina, the error in reaching each goal point was calculated using the following formula: $x_{error} = \|{ (x-x^{'}) }\|$,  $y_{error} = \|{ (y-y^{'}) }\|$ where $(x,y)$ is the desired point to reach in image coordinates and $(x^{'},y^{'})$ is the final landing position of the tool-tip in image coordinates. The image coordinates were converted into mm by finding the number of pixels of a known dimension. Specifically, the tool-tip measured 0.320mm and 8pixels in image coordinates. Thus, the pixel error values were converted to mm by multiplying the ratio of 0.320mm/8pixels.

\section{Eye Models}
\begin{figure}[h]
        \centering
        \includegraphics[width=1.0\textwidth]{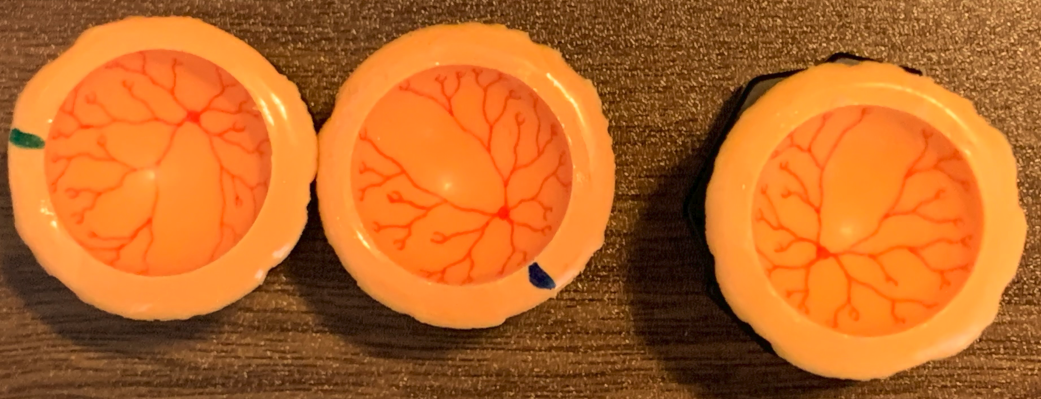}
        \caption{\small The two eyes on the left were used for training and the right eye was used for testing.}\vspace{-5pt}
        \label{fig:eye_models}
        \vspace{-5pt}
\end{figure}

As shown in Fig. \ref{fig:eye_models}, the two left eyes were used for training and the right eye was used for testing. We fabricated all the eyes by molding the appropriate polymer and drawing the blood-vessels using water-based paint. All eyes were 
molded using a 25.4 $\pm$ 0.05mm diameter sphere.

\section{Simulation Details}
\label{appendix_sim}
\begin{figure}[h]
        \centering
        \includegraphics[width=1.0\textwidth]{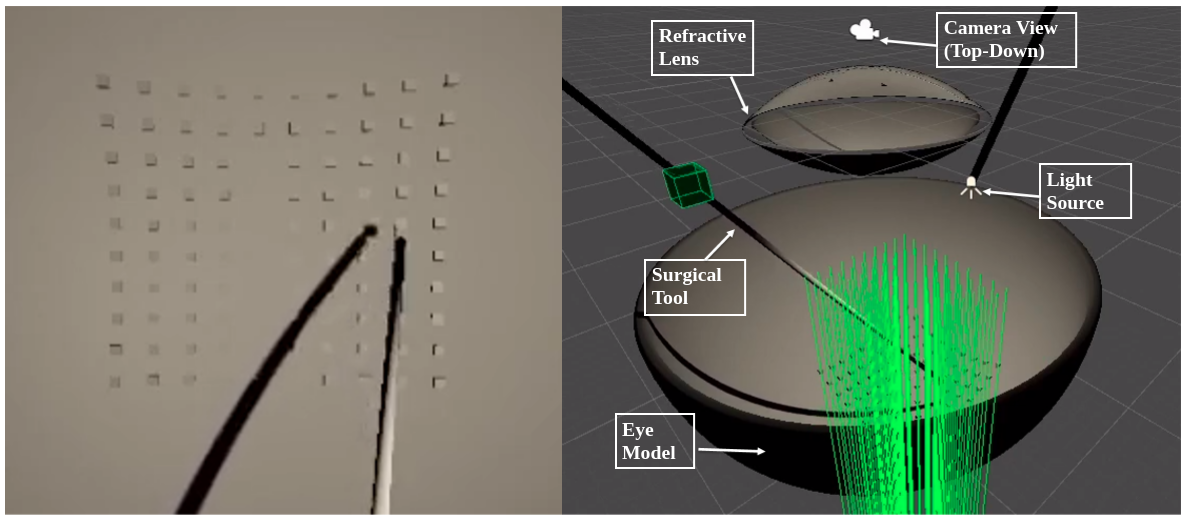}
        \caption{\small (Left) Benchmark task of visiting 100 array of positions in simulation; the center of each block represents each point to visit for visualization purpose. (Right) Raycast is used to find the ground-truth xyz points of the goal points to calculate errors accurately.}\vspace{-5pt}
        \label{fig:sim_details}
        \vspace{-5pt}
\end{figure}

As shown in Fig.\ref{fig:sim_details}, the benchmark task of visiting 100 array of points is demonstrated. In the left picture of Fig.\ref{fig:sim_details},  each goal-point-to-visit is represented as a block for visualization purpose. In the right image of Fig.\ref{fig:sim_details}, 100 raycasts are visualized, which are used to find the exact xyz locations of the retina at the desired goal positions for accurate error analysis. We also randomized the positions of the objects in the scene for robust generalization to changing objects in the scene: the eye was moved $\pm$0.5mm in xyz-axes. The  lens position was moved $\pm$0.5mm in  xyz-axes and rotated $\pm$3degrees in xyz-axes. The light source was also moved $\pm$1mm in xy-axes and $\pm$0.5mm in z-axis and its orientation was moved $\pm$1.5degrees along x-axis and $\pm$2.5 degrees along yz-axes.

\end{document}